\documentclass[letterpaper]{article} 
\usepackage{aaai2026}  
\usepackage{times}  
\usepackage{helvet}  
\usepackage{courier}  
\usepackage[hyphens]{url}  
\usepackage{graphicx} 
\usepackage{natbib}  
\usepackage{caption} 
\usepackage{amsmath}
\usepackage{amsfonts}
\usepackage{booktabs} 
\usepackage{multirow} 
\usepackage{graphicx}  
\usepackage{colortbl}
\usepackage{subcaption}
\usepackage{pifont}     
\usepackage{url}
\usepackage{graphicx}
\usepackage{amsthm}
\usepackage{multirow}
\usepackage{booktabs}

\usepackage{makecell}
\usepackage{pifont}
\usepackage{graphicx}
\usepackage{subcaption} 
\usepackage{algorithm}
\usepackage{algpseudocode}
\usepackage{changepage}
\usepackage{amssymb}
\usepackage{amsmath}
\usepackage{booktabs}
\usepackage{multirow}
\usepackage{float}
\theoremstyle{plain}

\newtheorem{theorem}{Theorem}

\newtheorem{lemma}[theorem]{Lemma}

\theoremstyle{definition}
\newtheorem{definition}[theorem]{Definition}

\theoremstyle{remark}

\usepackage{newfloat}
\usepackage{listings}
\DeclareCaptionStyle{ruled}{labelfont=normalfont,labelsep=colon,strut=off} 
\usepackage{float}
\floatstyle{ruled}
\newfloat{listing}{tb}{lst}{}
\floatname{listing}{Listing}
\title{Fairness-Aware Multi-view Evidential Learning with Adaptive Prior}
\author{
    Haishun Chen\textsuperscript{\rm 1},
    Cai Xu\textsuperscript{\rm 1},
    Jinlong Yu\textsuperscript{\rm 1},
    Yilin Zhang\textsuperscript{\rm 1},
    Ziyu Guan\textsuperscript{\rm 1},
    Wei Zhao\textsuperscript{\rm 1},\\
    Fangyuan Zhao\textsuperscript{\rm 2},
    Xin Yang\textsuperscript{\rm 1}
}
\affiliations{
    \textsuperscript{\rm 1}School of Computer Science and Technology, Xidian University, China\\
    \textsuperscript{\rm 2}School of Computer Science and Technology, Xi’an Jiaotong University, China\\
    \{chenhaishun@stu., cxu@, 24031212287@stu., ylzhang3@stu., zyguan@, ywzhao@mail., xdxinyang@stu.\}xidian.edu.cn\\
    zfy1454236335@stu.xjtu.edu.cn
}

\usepackage{bibentry}
\begin{document}

\maketitle

\begin{abstract}
Multi-view evidential learning aims to integrate information from multiple views to improve prediction performance and provide trustworthy uncertainty estimation. Most previous methods assume that view-specific evidence learning is naturally reliable. However, in practice, the evidence learning process tends to be biased. Through empirical analysis on real-world data, we reveal that samples tend to be assigned more evidence to support data-rich classes, thereby leading to unreliable uncertainty estimation in predictions. This motivates us to delve into a new Biased Evidential Multi-view Learning (BEML) problem. To this end, we propose Fairness-Aware Multi-view Evidential Learning (FAML). FAML first introduces an adaptive prior based on training trajectories, which acts as a regularization strategy to flexibly calibrate the biased evidence learning process. Furthermore, we explicitly incorporate a fairness constraint based on class-wise evidence variance to promote balanced evidence allocation. In the multi-view fusion stage, we propose an opinion alignment mechanism to mitigate view-specific bias across views, thereby encouraging the integration of consistent and mutually supportive evidence. Theoretical analysis shows that FAML enhances fairness in the evidence learning process. Extensive experiments on six real-world multi-view datasets demonstrate that FAML achieves more balanced evidence allocation and improves both prediction performance and the reliability of uncertainty estimation compared to state-of-the-art methods.
\end{abstract}


\section{Introduction}

Multi-view learning leverages complementary information from multiple views to improve model performance, providing significant advantages to real-world applications \cite{chen2025biased,sun2024robust}. For instance, autonomous driving systems integrate data from multiple sensors to accurately perceive the driving environment \cite{cheng2024adaptive}. However, in these high-stakes applications, it is essential for models to reliably represent the uncertainty associated with their predictions \cite{duan2024evidential}.

To address this practical limitation, a series of multi-view evidential learning (MVEL) methods based on Evidential Deep Learning (EDL)\cite{sensoy2018evidential} have been proposed to provide classification predictions alongside the corresponding uncertainty. In the prevailing MVEL paradigm, evidence is separately collected from each view in accordance with Subjective Logic (SL) \cite{josang2016subjective} and then aggregated to parameterize a Dirichlet distribution initialized with a non-informative uniform prior. Following this line of thought, many subsequent studies focus on refining the evidence-level fusion of multiple views to better handle real-world challenges such as inter-view conflict \cite{xu2024reliable,liu2024dynamic} and low-quality views \cite{deng2023uncertainty, xu2024trusted}.

\begin{figure}[t]
\centering
\includegraphics[width=0.95\columnwidth]{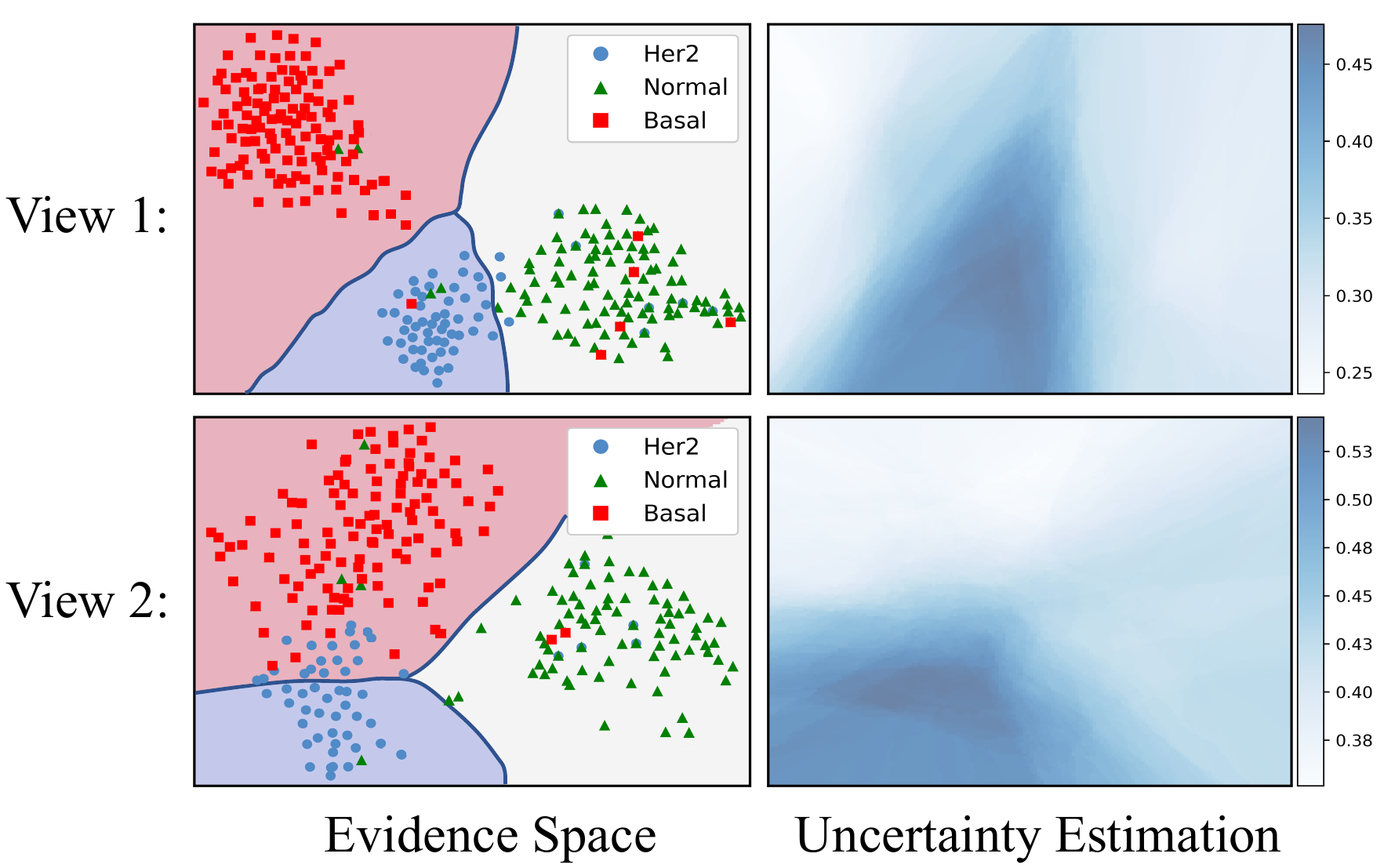} 
\caption{Visualization of biased evidence learning and uncertainty estimation on BRCA dataset. \textbf{Left:} Evidence space with training samples (dots) and categorical decision boundaries (in shade) for Her2, Normal, and Basal classes in two  views. \textbf{Right:} Corresponding uncertainty estimation, where darker regions indicate the prediction is more uncertain.}
\label{fig1}
\end{figure}

\begin{figure*}[t]
\centering
\includegraphics[width=0.99\textwidth]{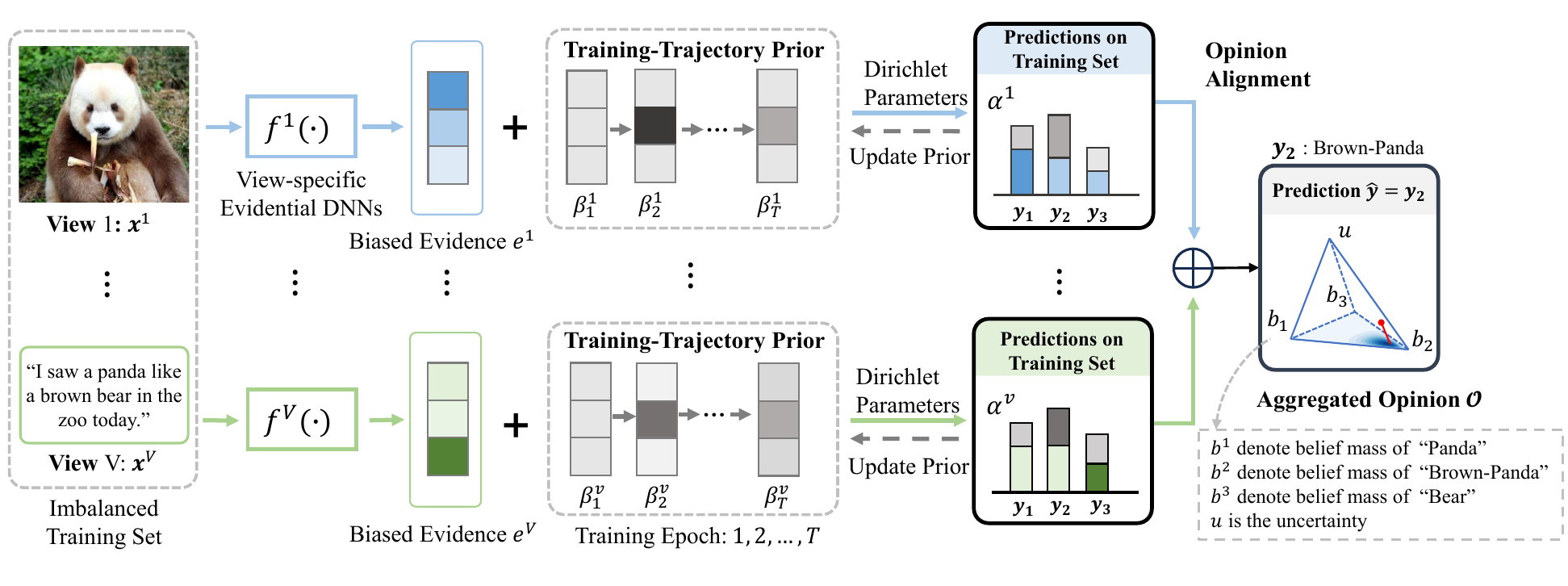} 
\caption{Illustration of FAML.We first employ view-specific evidential neural networks to construct opinions from each view. To mitigate evidential bias, a training-trajectory-based adaptive prior is introduced during evidence learning, ensuring fairer evidence allocation. An opinion alignment mechanism is then applied to promote consistency among different views. Finally, all view-specific opinions are integrated to reach a reliable and trustworthy decision.}
\label{fig2}
\end{figure*}

Although these methods achieve promising performance, most existing studies generally assume that view-specific evidence learning is inherently reliable. However, in practice, the evidence learning process tends to be biased, which ultimately leads to unfairness in the allocation of evidence. More intuitively, as shown in Figure 1, an empirical analysis is conducted on imbalanced real-world multi-view dataset Breast Invasive Carcinoma (BRCA) \cite{shea2020invasive}. We observe that for samples from the Her2 class, the supporting evidence is often largely assigned to the Normal class in the first view and to the Basal class in the second view. The view-specific biased evidence learning process leads Her2 samples to collect strong supporting evidence for data-rich classes, making them prone to being confidently misclassified. In contrast, correctly classified Her2 samples typically receive insufficient supporting evidence, which leads the model to assign low confidence to these correct predictions. This phenomenon poses a new Biased Evidential Multi-view Learning (BEML) problem, where the allocation of supporting evidence is inherently unfair, resulting in unreliable and biased predictive uncertainty.

To address the aforementioned problems, we propose a Fairness-Aware Multi-view Evidential Learning (FAML) method to address the BEML problem. Distinct from previous methods that primarily focus on how to aggregate evidence obtained from each view, FAML reformulates the evidential multi-view learning problem from the perspective of fair learning. As shown in Figure 2, to eliminate evidential bias across different classes, we introduce the training-trajectory-based prior into the construction of Dirichlet parameters, adaptively calibrating the prior support assigned to each class during training, thereby promoting balanced evidence allocation. In addition, we also calculate the fairness degree according to the variance of class-wise evidence as an explicit constraint to regularize the evidence learning process. Considering that bias exhibit view-specific pattern, we design an opinion alignment mechanism that minimizes the discrepancy of opinions between any pair of views, encouraging to gather consistent evidence from different views. 

The main contributions of this work are summarized as follows: (1) We reveal an neglected but widespread issue in multi-view evidential learning, where the evidence learning process often exhibits implicit unfairness in practice. (2) We propose an adaptive prior to resolve potential unfairness in multi-view evidential learning, and incorporate a fairness degree metric based on class-wise evidence variance to explicitly regularize evidence allocation during training. (3) Extensive experiments on five real-world multi-view datasets demonstrate that FAML consistently outperforms existing methods, achieving fair and reliable uncertainty estimation.

\section{Related work}
\subsubsection{Evidence Theory.} As an arising single-forward-pass uncertainty estimation method, evidential deep learning (EDL) has attracted increasing research attention in recent years. Grounded in the subjective logic theory, EDL models predictive outputs as evidence for a Dirichlet distribution, offering reliable uncertainty estimation in various downstream tasks \cite{gao2024comprehensive,xia2024uncertainty}. The effectiveness of EDL critically depends on the evidence collection process, with recent studies exploring improvements such as evidence allocation refinement \cite{li2024hyper}, sample reweighting \cite{pandey2023learn}, and advanced evidence functions\cite{shen2023post} to enhance uncertainty quantification. However, most recent works still rely on a fixed uniform prior for the Dirichlet distribution\cite{chen2024r}, which may undermine the reliability in uncertainty estimation in practice. To address this gap, we introduce a training-trajectory-based adaptive prior, explicitly designed to mitigate evidence bias and enhance uncertainty reliability.

\subsubsection{Multi-view Evidential Learning.} To enhance the performance and reliability of models by integrating data from multiple views, multi-view evidential learning has attracted significant attention. Pioneering works like Trusted Multi-view Classification \cite{han2022trusted} employ Dempster-Shafer theory to aggregate evidence from different views. Building on this research, a significant body of subsequent work has focused on making the fusion process more robust to real-world challenges. For example, some methods\cite{zhou2023calm,xu2024reliable} focus on explicitly handling conflicting opinions between different views to achieve a more reliable decision. Others\cite{liu2024dynamic,deng2025trustworthy} are designed to dynamically assign adaptive weights to each view, effectively reduce the influence of low-quality views during the fusion process. However, these methods focus on enhancing the consistency of evidence fusion across views, while the fairness within each view is of great significance for real-world high-stakes applications. Therefore, we propose to address the problem of unfair evidence learning to improve the reliability of the model.

\subsubsection{Fairness-aware Learning.}
With the growing attention to fairness-aware learning, recent studies have increasingly focused on the challenges posed by class imbalance. To alleviate such imbalance-induced unfairness, some approaches resort to data pre-processing, aiming to create a more balanced training set through resampling \cite{calmon2017optimized} or synthetic data generation \cite{iosifidis2018dealing}. Meanwhile, other works address fairness during model training by incorporating margin-based regularizers \cite{zafar2017fairness,iosifidis2019adafair}, or fairness-driven constraints into the loss function to promote more fair predictions\cite{krasanakis2018adaptive}. In addition, several post-processing strategies have been proposed, such as logits calibrationcite \cite{fish2016confidence} or prediction refinement\cite{iosifidis2019fae}, to further mitigate unfairness after the model has been trained. Nevertheless, in multi-view settings, different views may exhibit view-specific unfairness that conventional fairness-aware strategies fail to address, presenting new and unique challenges in achieving fair and reliable predictions.

\section{The Method}
In this section, we first define the BEML problem, then present our proposed method FAML in detail, together with the theoretical studies and discussion.

\subsection{Problem Formulation}
In the context of multi-view learning, each instance is characterized by multiple views. To clarity, considering a dataset $D=\{\left\{x_n^v\right\}_{v=1}^V, {y}_n\}_{n=1}^N$ with labels $y_n \in\{1,2, \cdots, K\}$ , where $x_n^{v} \in \mathbb{R}^{d_v}$ denotes the feature vector for the $v$-th view of the $n$-th instance. Multi-view evidential learning aims to integrate information from different views at the evidence level, and then aggregates the evidence to parameterize a Dirichlet distribution over class predictions, thereby enhancing both classification performance and reliability. However, in practice, the evidence learning process often exhibits quantity-induced bias, leading to an unfairness evidence allocation. To ensure the fairness in multi-view evidential learning, the expected evidence in each view assigned to the ground-truth class should be comparable.
\begin{equation}
\begin{split}
    \mathbb{E}_{(x, y=k) \in D}\left[e^v_k(x)\right] &\approx \mathbb{E}_{\left(x, y=k^{\prime}\right) \in D}\left[e^v_{k^{\prime}}(x)\right] \\
    & \hfill \forall k, k^{\prime} \in\{1,2, \ldots, K\}
\end{split}
\end{equation}

The goal of BEML is to eliminate evidence bias in learning process, and achieve fair evidence allocation and reliable uncertainty estimation.

\subsection{Fairness-aware Multi-view Evidence learning}
As shown in Figure 2, the overall framework is composed of view-specific evidential learning and evidential multi-view fusion stages. In the first stage, we employ evidential deep neural networks to extract view-specific evidence, which reflects the degree of support for each class from different views. To mitigate evidential bias in learning process, a training-trajectory-based adaptive prior is introduced to calibrate the initial support assigned to each class when parameterizing the Dirichlet distribution. Each view-specific Dirichlet distribution is then used to construct subjective opinions, characterized by a belief mass vector and uncertainty mass. To explicitly promote fairness, we compute a fairness degree for each view based on the variance of class-wise evidence and impose it as a regularization constraint during optimization. In the fusion stage, we introduce an opinion alignment mechanism that minimizes the discrepancy among view-specific opinions, encouraging the model to integrate consistent supportive evidence across views. Details will be elaborated as below.

\subsubsection{View-specific evidence learning.}
Evidential Deep Learning extends Subjective Logic (SL) to deep neural networks, providing the model with the ability to quantify uncertainty in its predictions. Specifically, given a specific view of an instance $x^v_n$, the evidence $\boldsymbol{e}=\left\{e_1, \ldots, e_k\right\}$ is obtained using Evidential Deep Neural Networks. The extracted evidence is then used to parameterize the Dirichlet distribution $\operatorname{Dir}(\boldsymbol{p}\mid\boldsymbol{\alpha})$, where the parameter of the Dirichlet distribution can be derived as (for clarify, we omit the super- and sub-scripts):
\begin{equation}
\alpha_k = e_k + 1
\end{equation}
where the constant serves as a parameter termed as an uniform Dirichlet prior, implicitly assuming each class is equally probable in the absence of evidence. According to Subjective Logic theory, the belief masses $b_k$ and the uncertainty mass $u$ are defined as follows:
\begin{equation}
b_k=\frac{e_k}{S}=\frac{\alpha_k-1}{S},\ \ u=1-\sum_{k=1}^K b_k=\frac{K}{S}
\end{equation}
where $S = \sum_{k=1}^K \alpha_k$ is defined as the Dirichlet strength. To generate the final predictions, Subjective Logic models each view-specific prediction as multinomial opinions $\mathcal{O} = (\boldsymbol{b},u)$, and the projected probability distribution of multinomial opinions is given by:
\begin{equation}
P_k=b_k+a_k \cdot u
\end{equation}
Normally, the base rate $a_k$ is to set the prior probabilities equal for each category.

While the above formulation enables uncertainty-aware multi-view prediction, it implicitly overlooks the impact of uniform Dirichlet prior assumptions on the fairness of evidence collection. In particular, an improper prior could dominate the Dirichlet posterior, especially for samples with less supporting evidence, undermining both the reliability and fairness of the multi-view learning process. To address this issue, we propose a training-trajectory-based prior as a form of adaptive regularization. Instead of employing an uniform Dirichlet prior in parameterizing the Dirichlet distribution, our approach adjusts the prior adaptively based on the learning history of each class. The training-trajectory-based prior for $k$-th class is defined as follows:
\begin{equation}
\begin{split}
\beta_k &= \gamma \cdot N_k / ( \sum_{n: y_n = k} \kappa(y_n, f_\theta(x_n)) ) \\
\text{where} \quad 
&\kappa\left(y_n, f_\theta(x_n)\right)= 
\begin{cases}
    1, & \text{if } y_n = f_\theta(x_n) \\
    0, & \text{if } y_n \neq f_\theta(x_n)
\end{cases}
\end{split} 
\end{equation}
where $N_k$ is the number of samples from $k$-th class, and $\gamma$ is a hyperparameter that controls the strength of the prior. To construct the trajectory-based prior, we train the model using empirical risk minimization and record the prediction results $f_{\theta_t}(x_n)$ for each sample in training epoch $t$ and then adaptively adjust the prior for each class based on its prediction. In this way, we update the Dirichlet concentration parameters for each class as:
\begin{equation}
    \hat\alpha_k=e_k+\beta_k
\end{equation}

This adaptive prior exploits the learning trajectory of each class to adaptively calibrate their contribution in the evidence learning process, thereby enhancing fairness and mitigating the evidential bias. To further get an intuitive sense of the level of fairness among classes, we introduce a measure named the fairness degree in Definition 1, which quantifies the variance of class-wise evidence allocation in each view.

\textbf{Definition: Fairness Degree}. Given the set of evidence vectors produced by Evidential Deep Neural Networks across all samples, the fairness degree is defined as the variance of the class-wise average evidence:
\begin{equation}
\begin{aligned}
\mathcal{FD}(\left\{e_n\right\}_{n=1}^N)  = \text{Var}(\{\bar{e}_k\}_{k=1}^K) =\frac{1}{K} \sum_{k=1}^K\left(\bar{e}_k-\bar{e}\right)^2
\end{aligned}
\end{equation}
where $\bar{e}_k$ represents the average evidence for $k$-th class, and $\bar{e}$ is the global average of these class-wise evidences.

Intuitively, this metric measures how biased the model is when allocating evidence. A low Fairness Degree $\mathcal{FD}$ is desirable, as it reflects that the model allocates similar amounts of evidence to all classes, indicating a more balanced and fair evidence assignment. In contrast, a larger $\mathcal{FD}$ reflects significant variation in evidence across classes, implying that the model’s evidence assignment is highly biased.

\subsubsection{Evidential Multi-view Fusion.}
In this subsection, we focus on multi-view fusion based on view-specific opinions. Since each view provides its own prediction and associated uncertainty, it is necessary to aggregate these opinions to reflect the overall uncertainty of the final decision. To obtain a more comprehensive joint opinion, we taking into account the view-specific uncertainty to calculate aggregated evidence. Rather than treating each view equally, we propose a weighted evidence aggregation method as follows:

\textbf{Definition: Weighted Evidence Aggregation.} Given two opinions $\mathcal{O}^A$ and $\mathcal{O}^B$ over an instance, let $e^A_k$, $e^B_k$ and $e_k$ represent the evidence for the $k$-th category from view $A$, view $B$, and the aggregated view, we have:
\begin{equation}
\begin{aligned}
e_k &= \frac{(1 - u^A)e_k^A + (1 - u^B)e_k^B}{2 - u^A - u^B} \\
&= \frac{c^A}{c^A + c^B} e_k^A + \frac{c^B}{c^A + c^B} e_k^B
\end{aligned}
\end{equation}
where $c^A = 1 - u^A$ and $c^B = 1 - u^B$ represent the confidence of view $A$ and view $B$. Following Definition 2, we combine the evidence from all views into a common aggregated representation. 

According to the Subjective Logic theory, we can get the final multi-view joint opinion, and thus get the final probability of each class and the overall uncertainty. To ensure different views to provide consistent evidence, we further introduce the Dissonance Degree of views in Definition 3, which is established according to opinion variance.

\textbf{Definition 3: Dissonance Degree of Views}. Given any view represented by the opinion $\mathcal{O}^v$, the variance of opinion for $k$-th class is derived from the Dirichlet PDF:
\begin{equation}
\begin{aligned}
\operatorname{Var}\left(\alpha_k^v\right) & =\frac{\alpha_k\left(\sum_{k=1}^K \alpha_k-\alpha_k\right)}{\left(\sum_{k=1}^K \alpha_k\right)^2\left(\sum_{k=1}^K \alpha_k+1\right)} \\
& =\frac{p_k^v\left(1-p_k^v\right) u^v}{K+u^v}
\end{aligned}
\end{equation}
then the dissonance degree between two views $A$ and $B$ is computed as:
\begin{equation}
\mathcal{O}_{\mathrm{con}}\left(\mathcal{O}^A, \mathcal{O}^B\right)=\sum_{k=1}^K\left|\operatorname{Var}\left(\alpha_k^A\right)-\operatorname{Var}\left(\alpha_k^B\right)\right|
\end{equation}

The Dissonance Degree provides a principled metric for quantifying the level of agreement in uncertainty estimation between any pair of views, thereby enabling regularization of opinion alignment in multi-view evidential learning.
\subsubsection{Loss Function.}
In this subsection, we will introduce the process of training evidential neural networks to obtain the multi-view joint opinion. Unlike the conventional approaches that produce point estimates of class probabilities via softmax, we employ the softmax layer with non-negative activation functions to collect evidence for each view. As previous discussed, to eliminate evidence bias, we introduce the adaptive trajectory-based prior to obtain the concentration parameters of the Dirichlet distribution.

For instance $\left\{x_n^{v}\right\}_{v=1}^V$, the supervised loss is derived from the expected cross-entropy between the true label and the Dirichlet mean, which can be induced as:
\begin{equation}
\begin{aligned}
L_{a c e}\left(\boldsymbol{\hat\alpha}_n\right) & =\int\left[\sum_{k=1}^K-y_{n k} \log p_{n k}\right] \frac{\sum_{k=1}^K p_{n k}^{\hat\alpha_{n k}-1}}{B\left(\boldsymbol{\hat\alpha}_n\right)} d \mathbf{p}_n \\
& =\sum_{j=1}^K y_{n k}\left(\psi\left(S_n\right)-\psi\left(\hat\alpha_{n k}\right)\right)
\end{aligned}
\end{equation}
where $\psi(\cdot)$ is the digamma function. 

However, the objective does not guarantee that the evidence generated is unbiased across classes. To address this issue, we introduce the fairness constraint into the loss function as follows:
\begin{equation}
\mathcal{L}_{acc}\left(\boldsymbol{\hat\alpha}_n\right)=\mathcal{L}_{ace}(\boldsymbol{\hat\alpha}_n) + \lambda \cdot \mathcal{FD}(\left\{e_n\right\}_{n=1}^N) 
\end{equation}
where $\lambda $ is a balancing coefficient gradually changing from 0 to 1 during training to control constraint strength. By increasing the influence of fairness constraint in loss, the fairness constraint is gradually enforced to fine-tune the evidence allocation, ensuring that the learned representations are not only accurate but also unbiased.

In order to ensure the consistency of results between different opinions during training, minimizing the degree of dissonance between opinions was adopted. The consistency loss for the instance $\left\{x_n^{v}\right\}_{v=1}^V$ as:

\begin{equation}
\mathcal{L}_{\text {con}}= \sum_{p=1}^V \sum_{p \neq q}^V  \mathcal{O}_{\mathrm{con}}\left(\mathcal{O}_n^p, \mathcal{O}_n^q\right)
\end{equation}

Finally, to counteract the bias introduced by the imbalanced data distribution, we construct the overall training objective based on a class-balanced learning strategy following \cite{xia2022hybrid}:
\begin{equation}
\mathcal{L}=\frac{1}{N_{y_n}} \left[\mathcal{L}_{acc}\left(\boldsymbol{\hat\alpha}_n\right)+\sum_{v=1}^V \mathcal{L}_{acc}\left(\boldsymbol{\hat\alpha}_n^v\right)\right] + \beta \cdot\mathcal{L}_{\text {con}}
\end{equation}

\section{Theoretical Studies}
In this section, we aim to establish the generalization guarantees of the FAML framework, particularly for minority classes. The analysis is grounded in the margin theory of statistical learning, which connects the margin distribution achieved on the training set to the model's generalization error. We first introduce the definition of \textit{Evidence Margin}.

\begin{definition}[Evidence Margin]
Let \( h_\theta : \mathcal{X} \to \mathbb{R}^K_+ \) denote a model, parameterized by \(\theta\), that maps a sample \(x_n\) to a \(K\)-dimensional non-negative evidence vector \( e_n = [e_{n1}, \ldots, e_{nK}] \). The \emph{evidence margin} \(\rho_n(\theta)\) of \(x_n\) is then defined as
\begin{equation}
\rho_n(\theta) = e_{nk} - \max_{j \neq k} \{e_{nj}\},
\end{equation}
where \(k\) is the ground-truth class label of \(x_n\).
\label{def:margin}
\end{definition}

To formally establish the connection between such a sample-wise margin and the model's overall generalization ability, we now introduce a foundational result from statistical learning theory. The following lemma provides a general bound that relates the margin distribution achieved on the training set to the upper bound of the generalization error.

\begin{lemma}[Margin Generalization Bound]
For a hypothesis space \(\mathcal{H}\), given any margin \(\rho > 0\), with probability at least \(1 - \delta\), for all \(h \in \mathcal{H}\), the true risk \(R(h)\) satisfies:
\begin{align}\label{equ:margi_bound}
R(h) \leq \hat{R}_{S,\rho}(h) + C_1 \sqrt{\frac{\text{Complexity}(\mathcal{H})}{N\rho^2}} + C_2 \sqrt{\frac{\ln(1/\delta)}{N}}
\end{align}
where \(\hat{R}_{S,\rho}(h)\) is the proportion of samples in the training set \(S\) with a margin less than \(\rho\), \(\text{Complexity}(\mathcal{H})\) is a complexity measure of the hypothesis space, and \(C_1, C_2\) are constants.
\label{lemma:lem_bound}
\end{lemma}
Lemma~\ref{lemma:lem_bound} shows that increasing the margin $\rho$ tightens the generalization bound via the $O(1/\rho)$ term. This allows us to use the change in $\rho$ as a bridge: by analyzing the effect of the FAML algorithm on $\rho$, we can explicitly determine its impact on the generalization error bound. 
\begin{theorem}
\label{thm:main-theorem}
Let \(h_{\theta}:\mathcal{X}\rightarrow\mathbb{R}_{+}^{K}\) be a model trained on a dataset \(S\) of size \(N\). For a minority class \(k\) with imbalance ratio \(\xi_{k}=N_{-k}/N_{k}\gg 1\), where \(N_k\) and \(N_{-k}\) are the number of samples belonging to class \(k\) and other classes respectively, and under the condition that the adaptive prior \(\beta_k\) increases when class \(k\) performs poorly, minimizing the FAML objective \(L(\theta)\) yields the following theoretical guarantees for the evidence margin \(\rho_{n}(\theta)=e_{nk}-\max_{j\neq k}e_{nj}\):

\begin{itemize}
    \item \textbf{Margin Increase for Minority Class.} The expected margin improvement for class \(k\) is bounded below by:
    \[
    \mathbb{E}[\Delta\rho_{n}\mid y_{n}=k]\gtrsim\eta\psi^{\prime}\left(S_{\hat{\alpha}}\right)\cdot\left(\xi_{k}\cdot\frac{\Delta\beta_{k}}{S_{\hat{\alpha}}}-O\left(\frac{1}{\hat{\alpha}_{nk}^{2}}\right)\right),
    \]
    where \(\eta\) is the learning rate, \(S_{\hat{\alpha}}=\sum_{j}\hat{\alpha}_{nj}\), \(\hat{\alpha}_{nj} = e_{nj} + \beta_j\), and \(\Delta\beta_k > 0\) is the increase in the adaptive prior. For \(\xi_{k}\gg 1\), the positive term dominates, ensuring a net margin increase.

    \item \textbf{Tighter Generalization Bound.} The increase in margin tightens the generalization bound. The effective margin \(\rho_{\textup{eff}}\) satisfies:
    \[
    \rho_{\textup{eff}}\gtrsim\rho_{0}\left(1+\lambda\cdot\xi_{k}\cdot\Delta\beta_{k}\right),
    \]
    where \(\rho_{0}\) is the baseline margin and \(\lambda>0\) is a scaling factor. This leads to an improved generalization error bound for the minority class by a factor of \(\tilde{O}\left(1/\sqrt{\xi_{k}\Delta\beta_{k}}\right)\).
\end{itemize}
\end{theorem}

Theorem~\ref{thm:main-theorem} delineates how the adaptive prior enhances the evidence margin and tightens the generalization bound, with particular efficacy for minority classes. First, as captured by the lower bound on the expected margin improvement, the corrective mechanism scales with the class imbalance ratio $\xi_k$. This scaling property ensures that the model automatically allocates greater corrective effort to classes that are both under-represented and under-performing. Second, the generalization improvement factor of $\tilde{O}(1/\sqrt{\xi_{k}\Delta\beta_{k}})$ quantifies this behavior: it shows that greater data imbalance (larger $\xi_k$) coupled with stronger model self-feedback (larger $\Delta\beta_k$) leads to greater generalization gain. This result provides a theoretical foundation for the superior performance of our method on long-tailed distributions.

\section{Experiment}
In this section, we conduct extensive experiments on five multi-view datasets. More implementation details are provided in Technical Appendix.
\begin{table*}[ht]
    \centering
    \renewcommand{\arraystretch}{0.6}
    \caption{Performance comparison in terms of ACC(\%) and ECE(\%) on test sets, ± indicates the standard deviation for 5 random seeds, with the best mean scores highlighted in \textbf{bold}.}
    \resizebox{\textwidth}{!}{%
    \begin{tabular}{c|c|cccc|ccccc} 
        \cmidrule(lr){1-10}
        \multirow{2}{*}{\textbf{Dataset}} & \multirow{2}{*}{\textbf{Method}} & \multicolumn{4}{c|}{\textbf{ACC(\%)} $\uparrow$} & \multicolumn{4}{c}{\textbf{ECE(\%)} $\downarrow$} \\
        & & All & Head & Med & Tail & All & Head & Med & Tail\\ \cmidrule(lr){1-10} 
        \multirow{8}{*}{Handwritten}
        & TLC  & 81.9 ± 1.1 & 98.0 ± 1.5  & 85.0 ± 1.9 & 76.5 ± 4.0 & 23.3 ± 1.3 & 38.1 ± 2.3 & 26.7 ± 1.8 & 22.6 ± 1.4 \\ 
        & I-EDL & 78.9 ± 1.7 & 93.0 ± 1.4 & 82.0 ± 2.5 & 65.9 ± 3.8  & 22.2 ± 3.3 & 26.8 ± 1.4 & 24.3 ± 1.5 & 32.2 ± 3.5\\
        & R-EDL & 85.5 ± 1.2 & 99.5 ± 0.4 & 88.6 ± 0.8 & 72.5 ± 2.5 & 34.9 ± 2.4 & 28.9 ± 0.5 & 31.5 ± 1.2 & 35.2 ± 3.9 \\
        & TMC & 84.2 ± 0.4 & 99.2 ± 0.2 & 89.0 ± 0.3 & 69.3 ± 1.0 & 28.9 ± 0.8 & 31.6 ± 0.5 & 23.1 ± 1.0 & 23.4 ± 3.5 \\
        & ETMC & 90.2 ± 0.8 & 99.2 ± 0.0 & 92.2 ± 0.4 & 83.1 ± 2.3 & 26.4 ± 1.2 & 19.6 ± 0.3 & 17.8 ± 1.0 & 24.2 ± 2.3 \\
        & CCML & 84.1 ± 2.4 & 99.0 ± 0.3  & 87.5 ± 1.7 & 70.2 ± 6.2 & 36.3 ± 1.8 & 34.2 ± 2.5 & 27.0 ± 1.4 & 36.3 ± 6.3\\ 
        & ECML & 79.4 ± 5.8 & \textbf{99.8 ± 0.3}  & 80.8 ± 1.1 & 63.4 ± 1.4 & 31.9 ± 5.4 & 40.9 ± 3.2 & 34.0 ± 2.7 & 21.0 ± 4.2\\  \cmidrule(lr){2-10} 
        & Ours & \textbf{94.2 ± 0.3} & 98.3 ± 0.4 & \textbf{92.5 ± 0.5} & \textbf{92.5 ± 0.7} & \textbf{20.6 ± 1.6} & \textbf{25.1 ± 2.4} & \textbf{17.0 ± 3.3} & \textbf{20.2 ± 4.3} \\ \cmidrule(lr){1-10} 
        \multirow{8}{*}{Animal} 
        & TLC  & 68.9 ± 1.8 & 86.3 ± 3.4 & 66.3 ± 7.0 & 55.6 ± 3.1 & 23.0 ± 5.0 & 37.3 ± 5.4 & 23.6 ± 3.8 & 21.6 ± 3.4 \\ 
        & I-EDL & 64.0 ± 1.1 & 91.6 ± 1.9 & 67.9 ± 2.4 & 36.0 ± 3.8 & 20.8 ± 2.6 & 47.6 ± 6.5 & 25.2 ± 4.5 & 22.9 ± 3.1 \\
        & R-EDL & 68.9 ± 1.7 & 90.0 ± 2.1 & 73.0 ± 4.4 & 46.4 ± 4.0 & 13.6 ± 1.2 & 36.0 ± 2.5 & 26.8 ± 2.3 & 23.2 ± 3.6 \\
        & TMC & 64.8 ± 0.3 & 89.6 ± 0.4 & 64.8 ± 0.4 & 42.7 ± 0.9 & 16.8 ± 0.4 & 37.4 ± 0.6 & 22.3 ± 1.7 & 20.1 ± 3.1 \\
        & ETMC & 64.5 ± 0.3 & 90.0 ± 0.4 & 64.8 ± 0.4 & 41.4 ± 0.8 & 12.2 ± 0.7 & 27.4 ± 0.6 & 18.7 ± 1.1 & 17.8 ± 1.0  \\
        & CCML & 64.6 ± 0.7 & 89.8 ± 0.7 & 63.8 ± 0.4 & 42.9 ± 1.1 & 23.1 ± 1.1 & 46.3 ± 1.2 & 23.3 ± 1.6 & 19.8 ± 2.1 \\
        & ECML & 65.5 ± 0.5 & 89.8 ± 0.7 & 65.3 ± 1.2 & 44.1 ± 0.3 & 20.5 ± 1.5 & 37.9 ± 1.1 & 23.9 ± 2.1 & 18.7 ± 1.9 \\
        \cmidrule(lr){2-10} 
        & Ours & \textbf{76.3 ± 0.4} & \textbf{91.6 ± 0.9} & \textbf{81.3 ± 0.7} & \textbf{57.9 ± 0.4} & \textbf{11.0 ± 1.0} & \textbf{13.7 ± 1.3} & \textbf{16.4 ± 1.7} & \textbf{17.6 ± 1.6} \\ \cmidrule(lr){1-10} 
        \multirow{8}{*}{Scene15}
        & TLC  & 38.7 ± 0.4 & 65.9 ± 1.7 & 29.8 ± 2.0 & 20.5 ± 0.9 & 19.3 ± 2.0 & 38.6 ± 3.2 & 17.6 ± 5.6 & 17.8 ± 1.0 \\
        & I-EDL& 25.6 ± 1.9 & 42.8 ± 5.9 & 18.6 ± 3.1 & 15.2 ± 3.0 & 16.6 ± 5.7 & 20.9 ± 5.6 & 21.3 ± 4.9 & 29.5 ± 4.2 \\
        & R-EDL& 44.3 ± 1.5 & 71.4 ± 5.7 & 30.8 ± 3.4 & 30.8 ± 4.5 & 31.2 ± 2.1 & 46.4 ± 7.2 & 28.8 ± 2.5 & 25.8 ± 4.6  \\
        & TMC & 42.0 ± 0.3 & 69.1 ± 1.3 & 30.5 ± 0.7 & 26.2 ± 1.2 & 11.8 ± 0.5 & 29.6 ± 1.5 & 13.7 ± 1.4 & 14.8 ± 1.8 \\
        & ETMC & 50.2 ± 0.4 & 73.3 ± 0.9 & 42.4 ± 0.5 & 34.8 ± 0.3 & 22.7 ± 0.3 & 32.6 ± 1.0 & 16.9 ± 1.0 & 21.1 ± 2.5  \\
        & CCML & 39.5 ± 0.9 & 63.9 ± 0.7 & 31.6 ± 0.7 & 23.1 ± 2.2 & 16.2 ± 0.5 & 28.6 ± 1.3 & 17.9 ± 1.0 & 17.5 ± 0.7 \\ 
        & ECML & 37.3 ± 0.7 & 65.2 ± 1.4 & 25.7 ± 0.9 & 20.9 ± 1.5 & 17.4 ± 0.9 & 35.9 ± 1.5 & 13.2 ± 1.1 & 18.0 ± 1.1 \\
        \cmidrule(lr){2-10} 
        & Ours & \textbf{57.2 ± 1.3} & \textbf{75.2 ± 1.7} & \textbf{50.1 ± 1.6} & \textbf{46.3 ± 3.9} & \textbf{11.2 ± 3.6} & \textbf{18.8 ± 2.5} & \textbf{13.2 ± 2.0} & \textbf{12.4 ± 1.1} \\ \cmidrule(lr){1-10} 
        \multirow{8}{*}{YaleB}
        & TLC  & 73.1 ± 4.7 & 89.5 ± 7.6  & 80.0 ± 4.6 & 55.7 ± 1.8 & 29.9 ± 6.5 & 30.9 ± 3.8 & 37.8 ± 3.8 & 46.3 ± 2.0\\ 
        & I-EDL & 83.7 ± 1.4 & 86.7 ± 3.5  & 82.9 ± 3.8 & 82.1 ± 5.0 & 35.2 ± 3.6 & 48.2 ± 5.4 & 32.1 ± 5.8 & 49.5 ± 4.9\\
        & R-EDL & 85.4 ± 1.0 & 96.2 ± 1.9 & 83.8 ± 3.8 & 78.6 ± 5.9 & 47.8 ± 4.2 & 37.3 ± 2.5 & 39.3 ± 1.5 & 45.9 ± 3.6 \\  
        & TMC  & 80.0 ± 0.6  & 98.1 ± 2.0  & 80.9 ± 0.4 & 64.3 ± 1.4 & 30.1 ± 2.1 & 52.9 ± 0.2 & 25.9 ± 0.7 & 39.9 ± 1.6\\ 
        & ETMC & 82.2 ± 0.4 & 99.1 ± 0.3 & 77.1 ± 1.9 & 72.9 ± 2.8 & 30.6 ± 2.4 & 55.7 ± 0.7 & 29.1 ± 3.5 & 38.8 ± 2.6  \\
        & CCML & 78.0 ± 4.7 & 98.1 ± 2.3  & 77.1 ± 1.9 & 63.6 ± 1.4 & 39.6 ± 3.5 & 54.3 ± 3.8 & 35.7 ± 4.8 & 38.9 ± 1.4\\  
        & ECML & 76.6 ± 1.1 & 90.5 ± 1.0  & 80.1 ± 2.1 & 62.9 ± 2.8 & 30.6 ± 2.8 & 51.1 ± 1.6 & 29.7 ± 1.9 & 45.1 ± 1.8\\
        \cmidrule(lr){2-10} 
        & Ours & \textbf{88.3 ± 2.9} & \textbf{99.5 ± 0.2} & \textbf{81.9 ± 1.0} & \textbf{84.3 ± 1.7} & \textbf{10.9 ± 3.4} & \textbf{27.2 ± 1.1} & \textbf{15.2 ± 1.3} & \textbf{27.5 ± 1.6} \\ \cmidrule(lr){1-10} 
        \multirow{8}{*}{Caltech-101}
        & TLC  & 74.2 ± 0.5 & 97.6 ± 0.9 & 72.9 ± 1.2 & 57.5 ± 1.3 & 26.2 ± 2.7 & 48.7 ± 3.0 & 29.6 ± 2.3 & 34.1 ± 3.1 \\ 
        & I-EDL & 75.9 ± 1.4 & 99.1 ± 1.9 & 82.4 ± 1.1 & 53.6 ± 2.8 & 32.3 ± 2.6 & 56.9 ± 4.3 & 37.4 ± 3.7 & 30.5 ± 2.9 \\ 
        & R-EDL & 70.0 ± 2.9 & 95.7 ± 1.7 & 77.6 ± 2.4 & 45.0 ± 3.7 & 24.8 ± 3.5 & 23.9 ± 4.4 & 34.7 ± 4.5 & 31.5 ± 3.8 \\
        & TMC & 73.9 ± 2.2 & 95.2 ± 2.1 & 79.5 ± 4.4 & 53.6 ± 2.9 & 19.8 ± 2.3 & 25.8 ± 3.9 & 33.4 ± 3.9 & 31.9 ± 1.4 \\
        & ETMC & 72.9 ± 1.1 & 95.2 ± 1.7 & 76.2 ± 3.2 & 53.6 ± 2.1 & 15.2 ± 2.4 & 21.1 ± 0.7 & 29.5 ± 3.5 & 30.4 ± 2.6  \\
        & CCML & 71.9 ± 1.2 & 97.6 ± 2.1 & 71.9 ± 3.1 & 52.5 ± 1.8 & 29.4 ± 0.9 & 45.1 ± 2.6 & 31.9 ± 3.0 & 32.7 ± 1.9 \\
        & ECML & 74.7 ± 0.8 & 97.1 ± 0.9 & 76.7 ± 6.6 & 56.4 ± 2.9 & 36.4 ± 1.7 & 45.2 ± 3.6 & 42.2 ± 4.1 & 34.3 ± 3.8 \\
        \cmidrule(lr){2-10} 
        & Ours & \textbf{83.6 ± 1.0} & \textbf{99.5 ± 0.9} & \textbf{88.1 ± 1.5} & \textbf{67.8 ± 2.8} & \textbf{14.1 ± 1.3} & \textbf{11.4 ± 4.2} & \textbf{17.2 ± 1.5} & \textbf{26.9 ± 3.4}  \\ \cmidrule(lr){1-10} 
    \end{tabular}
    }
    \label{performance results}
\end{table*}
\subsection{Experimental Setup}
\subsubsection{Dataset.}
To validate the effectiveness of FAML, we conduct experiments on five widely used real-word multi-view datasets as follows: \textbf{Handwritten}\footnote{https://archive.ics.uci.edu/dataset/72/multiple+features} is a six-view dataset, which comprises 2,000 instances of handwritten numerals ranging from ’0’ to ’9’. \textbf{Animal}\footnote{ https://www.vision.caltech.edu/Image} contains 2 views and 10,158 images from 50 categories, utilizing two types of deep features extracted using DEACA and VGG. \textbf{Scene15}\footnote{https://doi.org/10.6084/m9.figshare.7007177.v1} includes 4485 images from 15 indoor and outdoor scene categories. We extract three types of features GIST, PHOG, and LBP. \textbf{YaleB}\footnote{https://vision.ucsd.edu/content/yale-face-database} consists of 650 facial images from 10 categories with three views \textbf{Caltech101}\footnote{https://data.caltech.edu/records/mzrjq-6wc02} comprises 8677 images from 101 classes. We select 2386 samples from 20 classes. We extract Gabor, Wavelet Moments, CENTRIST, HOG, GIST, and LBP as 6 views. 
\subsubsection{Compared Methods.} We compare the proposed method with the following methods. (1) Single-view evidential learning methods contain: \textbf{TLC} (Trustworthy Long-tailed Classification) \cite{li2022trustworthy} combines class-imbalance classification and uncertainty estimation in a multi-expert framework. \textbf{I-EDL}( Information-based EDL) \cite{deng2023uncertainty} is the SOTA evidential approach for uncertainty-aware learning in the single-view setting. \textbf{R-EDL} (Relaxed-EDL) \cite{chen2024r} explores the relaxation of subjective logic assumptions to improve robustness. For single-view baselines, we report the results from the best-performing view. (2) Multi-view evidential learning methods contain: \textbf{TMC} (Trusted Multi-view Classifiction) \cite{han2022trusted} addresses the uncertainty estimation problem and produces reliable classification results. \textbf{CCML} (Consistent and Complementary-aware trusted Multi-view Learning) \cite{liu2024dynamic} explicitly decouples two types of evidence to enhance robustness to ambiguous views. \textbf{ECML}(Evidential Conflictive Multi-view Learning) \cite{xu2024reliable} is the SOTA method that proposed a fusion strategy for solving multi-view conflictive problems.
\subsubsection{Implementation Details.}
We implement all experiments on the PyTorch 1.13.1 framework. In our model, the view-specific evidence is extracted by fully connected networks with a ReLU layer. The model is trained for 200 epochs using the Adam optimizer with L2-norm regularization. More specifically, to avoid the influence of inaccurate predictions at the beginning of training, we adjust the prior with predictions after training for 20 epochs and then update it every five epochs to ensure training stability. In all dataset, we run 5 times for each compared method to report the mean value and standard deviations.

In our setting, the training data exhibits significant class imbalance, where the classes can be equally separated into head, medium and tail regions based on the different numbers of samples. Following the setting in most existing works on class imbalance problem, the test set is constructed to be class-balanced. Specifically, we randomly select 20\% of the samples as the testing set and use the remaining data to construct a revised version of the training set by sampling a subset that follows a Pareto distribution. 


\subsection{Experiment Results}
\subsubsection{Performance Comparison.} We evaluate the performances on classification task in terms of accuracy (ACC) on five real-world multi-view datasets. The results reveal several important observations: (1) FAML achieves superior performance compared to all other methods. While some methods, such as ECML, occasionally outperform FAML in isolated cases, FAML demonstrates more stable and consistent performance across all regions, especially in tail classes regions. (2) The fusion of multiple views does not guarantee better performance, as the aggregation of biased evidence across views can result in degraded predictive accuracy. 

We also use the Expected Calibration Error (ECE) to quantify the uncertainty estimation for the compared methods, which measure the correspondence between predicted probabilities and empirical accuracy. The results in terms of ECE reveal that our method significantly outperforms all compared methods, consistently yielding the lowest expected calibration error. This indicate FAML provides more precise uncertainty estimates with lower calibration errors.
\subsubsection{Ablation Studies.} To validate the effectiveness of adaptive prior, fairness constraint and consistency regularization, we construct a detailed ablation study on Caltech-101 dataset that performs different combinations of these modules to achieve degradation methods. The results are listed in Table 2. It is easy to conclude: (1) After removing any of proposed modules, the results show significant performance degradation, which shows that all components in FAML are indispensable. (2) Compared to the baseline with the three modules removed, we can observe the outstanding effectiveness of our adaptive prior strategy. (3) Compared to the consistency regularization, the fairness constraint provides a larger performance improvements.

\begin{table}[ht]
  \centering
  \renewcommand{\arraystretch}{0.8}
  \label{tab:ablation}
  \resizebox{\linewidth}{!}{ 
  \begin{tabular}{ccc|c}
    \toprule
    Adaptive Prior & Fairness & Consistency & ACC(\%) \\
    \midrule
    - & - & - & $73.8 \pm 2.2$ \\
    \checkmark & - & - & $ 79.1 \pm 0.5$ \\
    \checkmark & \checkmark & - & $84.3 \pm 0.2$ \\
    \checkmark & - & \checkmark & $80.0 \pm 0.9$ \\
    \checkmark & \checkmark & \checkmark & $87.1 \pm 0.9$ \\
    \bottomrule
  \end{tabular}
  }
  \caption{Ablation study on Caltech-101 dataset. "\checkmark" means FAML with the corresponding component, "-" means not applied.}
\end{table}
\begin{figure}[t]
  \centering
  \begin{subfigure}{0.49\linewidth}
    \includegraphics[width=\linewidth]{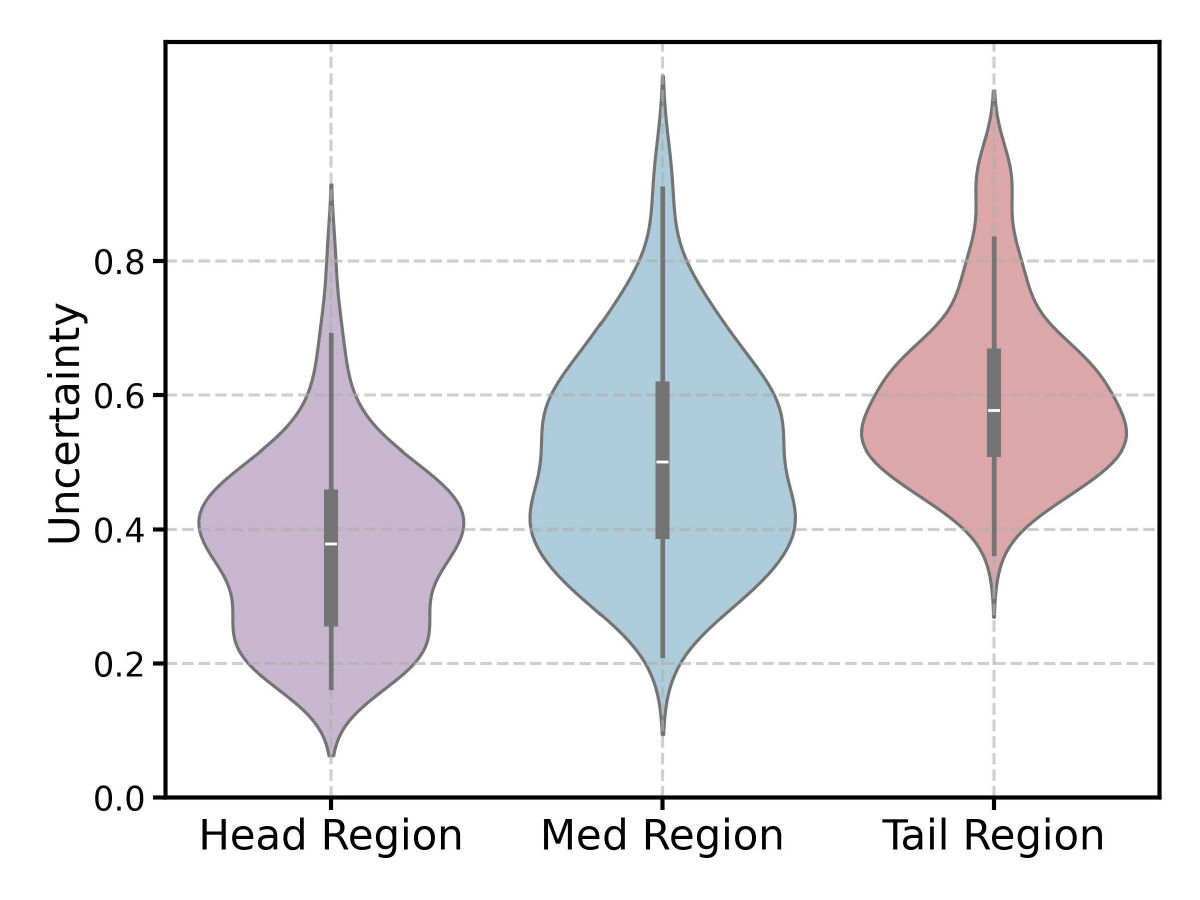}
    \caption{TMC}
    \label{fig:fig1}
  \end{subfigure}
  \hfill
  \begin{subfigure}{0.49\linewidth}
    \includegraphics[width=\linewidth]{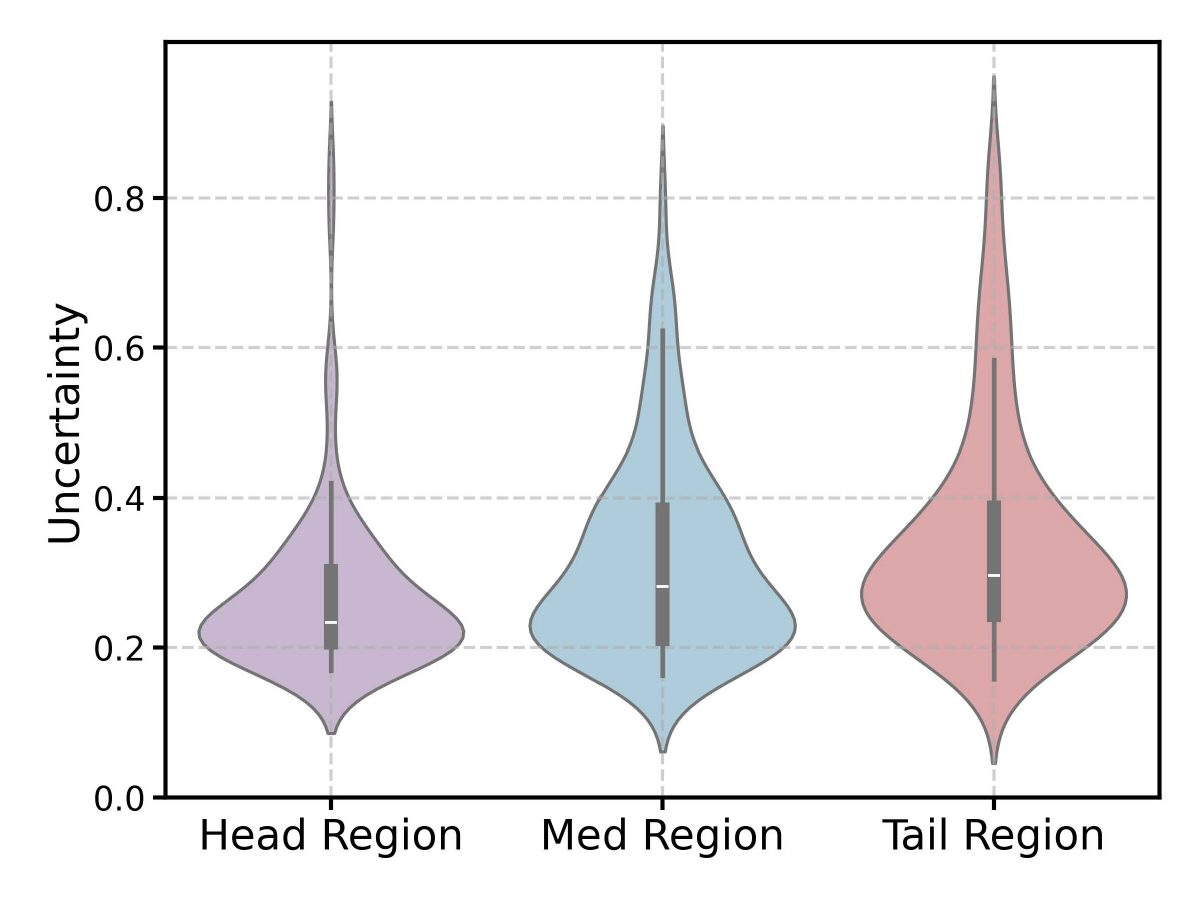}
    \caption{Ours}
    \label{fig:fig2}
  \end{subfigure}
  \caption{Uncertainty Estimation on Animal dataset.}
  \label{fig:Visualization}
\end{figure}

\subsubsection{Uncertainty Estimation.} To gain a deeper understanding of our proposed method, we visualize the uncertainty estimation distribution on the Animal dataset. As shown in Figure 3, we reveal the following observations: (1) For the TMC method, the uncertainty estimates are relatively low for head classes, but increase substantially for medium and tail classes. The samples from data-rich classes tend to be assigned low uncertainty due to the evidence collection process is biased, suggesting an unfair distribution of supporting evidence across classes. (2) In contrast, our method consistently produces more uniform and lower uncertainty. This indicates FAML effectively mitigates evidence allocation bias and delivers trustworthy predictions, demonstrating the fairness of FAML in estimating uncertainty.

\subsubsection{Visualization of Evidence Strength.}
In real-world datasets,  samples from data-rich classes are more likely to receive abundant supporting evidence, while those from other classes often receive much less. To further illustrate evidence allocation, Figure 4 shows the average evidence strength of each category. We observe that TMC heavily tend to collect more evidence in data-rich classes while much less evidence in data-poor classes. In contrast, our method maintains a more balanced evidence strength across all classes, demonstrating that FAML effectively alleviates evidence allocation bias and promotes fairness.

\begin{figure}[t]
  \centering
  \begin{subfigure}{0.49\linewidth}
    \includegraphics[width=\linewidth]{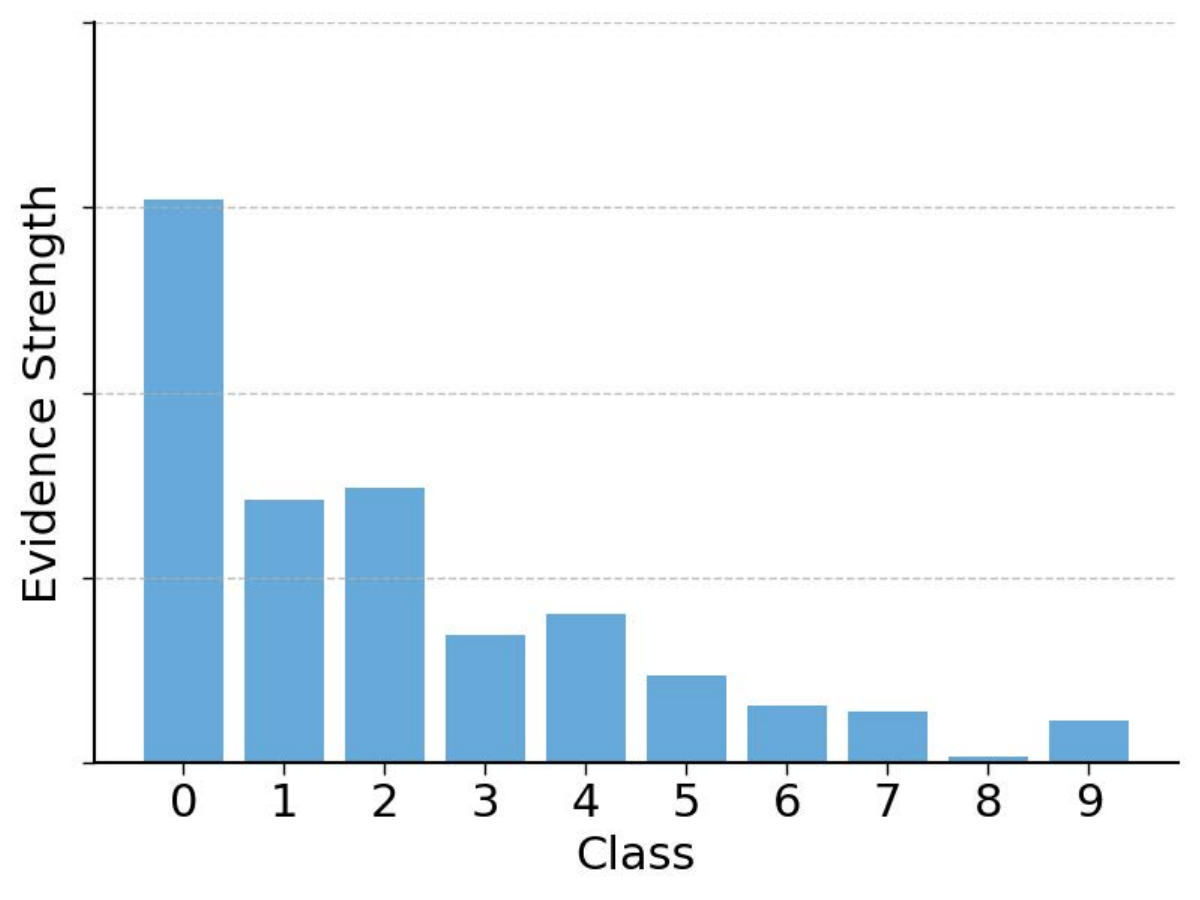}
    \caption{TMC}
    \label{fig:fig1}
  \end{subfigure}
  \hfill
  \begin{subfigure}{0.49\linewidth}
    \includegraphics[width=\linewidth]{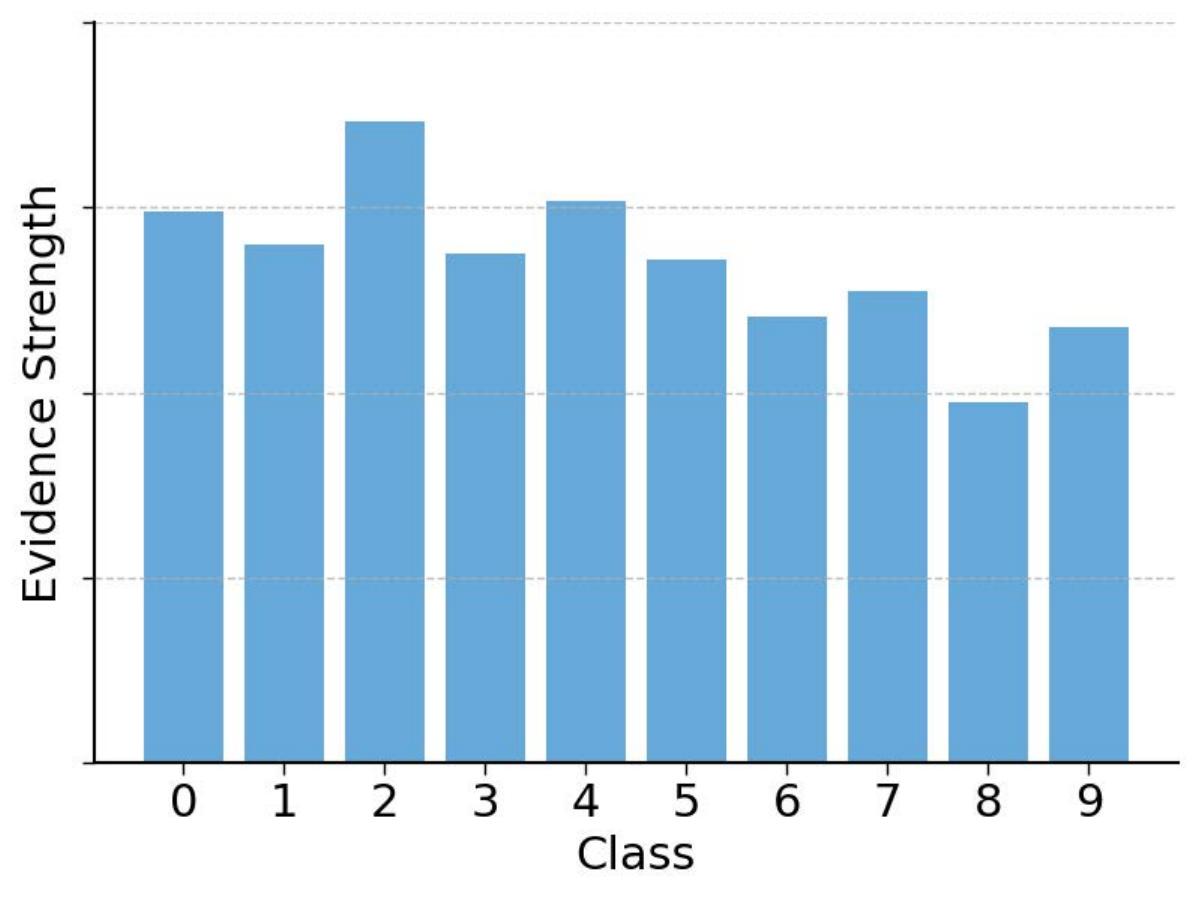}
    \caption{Ours}
    \label{fig:fig2}
  \end{subfigure}
  \caption{The average evidence strength of each category on Handwritten dataset.}
  \label{fig:Visualization}
\end{figure}

\subsubsection{Parameter Analysis.} To investigate the influence hyperparameter  $\gamma$, we perform parameter sensitivity analysis on across five datasets, with $\gamma$ ranging from 0.1 to 10.0 in our experiments. As shown in Figure 5, the optimal value of $\gamma$ varies across datasets, yet the best performance is consistently achieved within the range of 1.0 to 10.0. This suggests amplifying the differences among priors is beneficial for training, as it enables more effective utilization of priors. Taking this into consideration, we have determined the appropriate value of $\gamma$ for the evaluation experiments.

\begin{figure}[t]
\centering
\includegraphics[width=0.9\columnwidth]{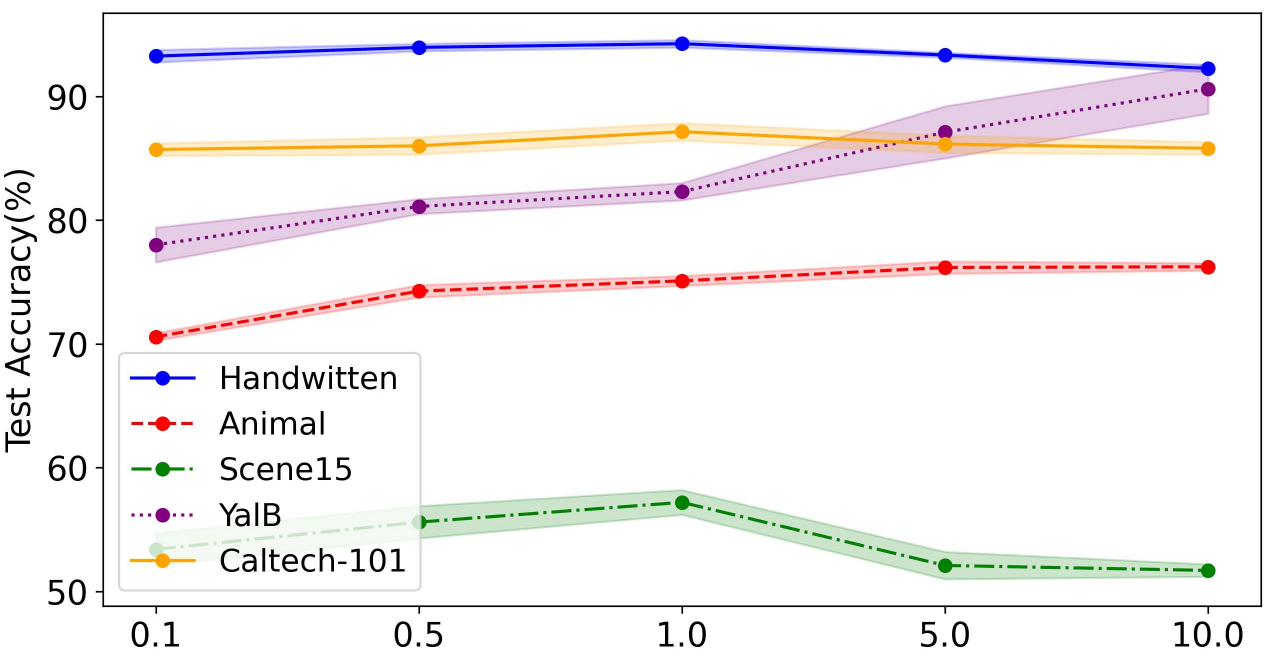} 
\caption{Accuracy(\%) when adjusting $\gamma$ on datasets.}
\label{fig6}
\end{figure}

\section{Conclusion}
In this paper, we introduced a Fairness-aware multi-view evidential Learning (FAML) method for addressing the biased evidetial multi-view learning problem. FAML introduces a training-trajectory-based adaptive prior instead of non-informative uniform prior to construct Dirichlet distribution, which adaptively calibrates the Dirichlet parameters to mitigate view-specific evidenctial bias. Futhermore,  we incorporate a fairness degree to constrain the view-speicific evidence learning process, thus explicitly enhancing balanced evidence allocation. During the fusion stage, we incorporate an opinion alignment mechanism to guide the formation of the final fused opinion. Extensive experiments on real-world datasets confirm the effectiveness of FAML on prediction performance and reliable uncertainty estimation.

\bibliography{main}

\begin{thebibliography}{29}
\providecommand{\natexlab}[1]{#1}

\bibitem[{Calmon et~al.(2017)Calmon, Wei, Vinzamuri, Natesan~Ramamurthy, and Varshney}]{calmon2017optimized}
Calmon, F.; Wei, D.; Vinzamuri, B.; Natesan~Ramamurthy, K.; and Varshney, K.~R. 2017.
\newblock Optimized pre-processing for discrimination prevention.
\newblock \emph{Advances in neural information processing systems}, 30.

\bibitem[{Chen et~al.(2025)Chen, Xu, Guan, Zhao, and Liu}]{chen2025biased}
Chen, H.; Xu, C.; Guan, Z.; Zhao, W.; and Liu, J. 2025.
\newblock Biased Incomplete Multi-View Learning.
\newblock In \emph{Proceedings of the AAAI Conference on Artificial Intelligence}, volume~39, 15767--15775.

\bibitem[{Chen, Gao, and Xu(2024)}]{chen2024r}
Chen, M.; Gao, J.; and Xu, C. 2024.
\newblock R-edl: Relaxing nonessential settings of evidential deep learning.
\newblock In \emph{The Twelfth International Conference on Learning Representations}.

\bibitem[{Cheng et~al.(2024)Cheng, Yin, Wang, Chen, Wang, and Yang}]{cheng2024adaptive}
Cheng, J.; Yin, W.; Wang, K.; Chen, X.; Wang, S.; and Yang, X. 2024.
\newblock Adaptive fusion of single-view and multi-view depth for autonomous driving.
\newblock In \emph{Proceedings of the IEEE/CVF Conference on Computer Vision and Pattern Recognition}, 10138--10147.

\bibitem[{Deng et~al.(2023)Deng, Chen, Yu, Liu, and Heng}]{deng2023uncertainty}
Deng, D.; Chen, G.; Yu, Y.; Liu, F.; and Heng, P.-A. 2023.
\newblock Uncertainty estimation by fisher information-based evidential deep learning.
\newblock In \emph{International conference on machine learning}, 7596--7616. PMLR.

\bibitem[{Deng et~al.(2025)Deng, Pan, Tang, and Shi}]{deng2025trustworthy}
Deng, H.; Pan, N.; Tang, C.; and Shi, L. 2025.
\newblock Trustworthy data recovery for incomplete multi-view learning.
\newblock \emph{Signal Processing}, 110146.

\bibitem[{Duan et~al.(2024)Duan, Caffo, Bai, Sair, and Jones}]{duan2024evidential}
Duan, R.; Caffo, B.; Bai, H.~X.; Sair, H.~I.; and Jones, C. 2024.
\newblock Evidential uncertainty quantification: A variance-based perspective.
\newblock In \emph{Proceedings of the IEEE/CVF Winter Conference on Applications of Computer Vision}, 2132--2141.

\bibitem[{Fish, Kun, and Lelkes(2016)}]{fish2016confidence}
Fish, B.; Kun, J.; and Lelkes, {\'A}.~D. 2016.
\newblock A confidence-based approach for balancing fairness and accuracy.
\newblock In \emph{Proceedings of the 2016 SIAM international conference on data mining}, 144--152. SIAM.

\bibitem[{Gao et~al.(2024)Gao, Chen, Xiang, and Xu}]{gao2024comprehensive}
Gao, J.; Chen, M.; Xiang, L.; and Xu, C. 2024.
\newblock A comprehensive survey on evidential deep learning and its applications.
\newblock \emph{arXiv preprint arXiv:2409.04720}.

\bibitem[{Han et~al.(2022)Han, Zhang, Fu, and Zhou}]{han2022trusted}
Han, Z.; Zhang, C.; Fu, H.; and Zhou, J.~T. 2022.
\newblock Trusted multi-view classification with dynamic evidential fusion.
\newblock \emph{IEEE transactions on pattern analysis and machine intelligence}, 45(2): 2551--2566.

\bibitem[{Iosifidis, Fetahu, and Ntoutsi(2019)}]{iosifidis2019fae}
Iosifidis, V.; Fetahu, B.; and Ntoutsi, E. 2019.
\newblock Fae: A fairness-aware ensemble framework.
\newblock In \emph{2019 IEEE international conference on big data (big data)}, 1375--1380. IEEE.

\bibitem[{Iosifidis and Ntoutsi(2018)}]{iosifidis2018dealing}
Iosifidis, V.; and Ntoutsi, E. 2018.
\newblock Dealing with bias via data augmentation in supervised learning scenarios.
\newblock \emph{Jo Bates Paul D. Clough Robert J{\"a}schke}, 24(11).

\bibitem[{Iosifidis and Ntoutsi(2019)}]{iosifidis2019adafair}
Iosifidis, V.; and Ntoutsi, E. 2019.
\newblock Adafair: Cumulative fairness adaptive boosting.
\newblock In \emph{Proceedings of the 28th ACM international conference on information and knowledge management}, 781--790.

\bibitem[{J{\o}sang(2016)}]{josang2016subjective}
J{\o}sang, A. 2016.
\newblock \emph{Subjective logic}, volume~3.
\newblock Springer.

\bibitem[{Krasanakis et~al.(2018)Krasanakis, Spyromitros-Xioufis, Papadopoulos, and Kompatsiaris}]{krasanakis2018adaptive}
Krasanakis, E.; Spyromitros-Xioufis, E.; Papadopoulos, S.; and Kompatsiaris, Y. 2018.
\newblock Adaptive sensitive reweighting to mitigate bias in fairness-aware classification.
\newblock In \emph{Proceedings of the 2018 world wide web conference}, 853--862.

\bibitem[{Li et~al.(2022)Li, Han, Li, Fu, and Zhang}]{li2022trustworthy}
Li, B.; Han, Z.; Li, H.; Fu, H.; and Zhang, C. 2022.
\newblock Trustworthy long-tailed classification.
\newblock In \emph{Proceedings of the IEEE/CVF Conference on Computer Vision and Pattern Recognition}, 6970--6979.

\bibitem[{Li et~al.(2024)Li, Li, Ou, Kaplan, J{\o}sang, Cho, Jeong, and Chen}]{li2024hyper}
Li, C.; Li, K.; Ou, Y.; Kaplan, L.~M.; J{\o}sang, A.; Cho, J.-H.; Jeong, D.~H.; and Chen, F. 2024.
\newblock Hyper evidential deep learning to quantify composite classification uncertainty.
\newblock \emph{arXiv preprint arXiv:2404.10980}.

\bibitem[{Liu et~al.(2024)Liu, Liu, Xu, Song, Guan, and Zhao}]{liu2024dynamic}
Liu, Y.; Liu, L.; Xu, C.; Song, X.; Guan, Z.; and Zhao, W. 2024.
\newblock Dynamic evidence decoupling for trusted multi-view learning.
\newblock In \emph{Proceedings of the 32nd ACM International Conference on Multimedia}, 7269--7277.

\bibitem[{Pandey and Yu(2023)}]{pandey2023learn}
Pandey, D.~S.; and Yu, Q. 2023.
\newblock Learn to accumulate evidence from all training samples: theory and practice.
\newblock In \emph{International Conference on Machine Learning}, 26963--26989. PMLR.

\bibitem[{Sensoy, Kaplan, and Kandemir(2018)}]{sensoy2018evidential}
Sensoy, M.; Kaplan, L.; and Kandemir, M. 2018.
\newblock Evidential deep learning to quantify classification uncertainty.
\newblock \emph{Advances in neural information processing systems}, 31.

\bibitem[{Shea, Koh, and Tan(2020)}]{shea2020invasive}
Shea, E. K.~H.; Koh, V. C.~Y.; and Tan, P.~H. 2020.
\newblock Invasive breast cancer: Current perspectives and emerging views.
\newblock \emph{Pathology international}, 70(5): 242--252.

\bibitem[{Shen et~al.(2023)Shen, Bu, Sattigeri, Ghosh, Das, and Wornell}]{shen2023post}
Shen, M.; Bu, Y.; Sattigeri, P.; Ghosh, S.; Das, S.; and Wornell, G. 2023.
\newblock Post-hoc uncertainty learning using a dirichlet meta-model.
\newblock In \emph{Proceedings of the AAAI Conference on Artificial Intelligence}, volume~37, 9772--9781.

\bibitem[{Sun et~al.(2024)Sun, Qin, Li, Peng, Peng, and Hu}]{sun2024robust}
Sun, Y.; Qin, Y.; Li, Y.; Peng, D.; Peng, X.; and Hu, P. 2024.
\newblock Robust multi-view clustering with noisy correspondence.
\newblock \emph{IEEE Transactions on Knowledge and Data Engineering}.

\bibitem[{Xia et~al.(2024)Xia, Dang, Han, Qendro, and Mascolo}]{xia2024uncertainty}
Xia, T.; Dang, T.; Han, J.; Qendro, L.; and Mascolo, C. 2024.
\newblock Uncertainty-aware health diagnostics via class-balanced evidential deep learning.
\newblock \emph{IEEE Journal of Biomedical and Health Informatics}, 28(11): 6417--6428.

\bibitem[{Xia et~al.(2022)Xia, Han, Qendro, Dang, and Mascolo}]{xia2022hybrid}
Xia, T.; Han, J.; Qendro, L.; Dang, T.; and Mascolo, C. 2022.
\newblock Hybrid-edl: Improving evidential deep learning for uncertainty quantification on imbalanced data.
\newblock In \emph{Workshop on Trustworthy and Socially Responsible Machine Learning, NeurIPS 2022}.

\bibitem[{Xu et~al.(2024{\natexlab{a}})Xu, Si, Guan, Zhao, Wu, and Gao}]{xu2024reliable}
Xu, C.; Si, J.; Guan, Z.; Zhao, W.; Wu, Y.; and Gao, X. 2024{\natexlab{a}}.
\newblock Reliable conflictive multi-view learning.
\newblock In \emph{Proceedings of the AAAI conference on artificial intelligence}, volume~38, 16129--16137.

\bibitem[{Xu et~al.(2024{\natexlab{b}})Xu, Zhang, Guan, and Zhao}]{xu2024trusted}
Xu, C.; Zhang, Y.; Guan, Z.; and Zhao, W. 2024{\natexlab{b}}.
\newblock Trusted multi-view learning with label noise.
\newblock \emph{arXiv preprint arXiv:2404.11944}.

\bibitem[{Zafar et~al.(2017)Zafar, Valera, Gomez~Rodriguez, and Gummadi}]{zafar2017fairness}
Zafar, M.~B.; Valera, I.; Gomez~Rodriguez, M.; and Gummadi, K.~P. 2017.
\newblock Fairness beyond disparate treatment \& disparate impact: Learning classification without disparate mistreatment.
\newblock In \emph{Proceedings of the 26th international conference on world wide web}, 1171--1180.

\bibitem[{Zhou et~al.(2023)Zhou, Xue, Liu, Li, Du, Liang, and Qi}]{zhou2023calm}
Zhou, H.; Xue, Z.; Liu, Y.; Li, B.; Du, J.; Liang, M.; and Qi, Y. 2023.
\newblock Calm: An enhanced encoding and confidence evaluating framework for trustworthy multi-view learning.
\newblock In \emph{Proceedings of the 31st ACM International Conference on Multimedia}, 3108--3116.

\end{thebibliography}
\end{document}